\pdfoutput=1

\documentclass[11pt]{article}

\usepackage[final]{acl}

\usepackage{times}
\usepackage{latexsym}
\usepackage{booktabs}
\usepackage{multirow}
\usepackage{subcaption}
\usepackage[T1]{fontenc}

\usepackage[utf8]{inputenc}

\usepackage{microtype}

\usepackage{inconsolata}
\usepackage{booktabs}
\usepackage{adjustbox}
\usepackage{graphicx}
\usepackage{listings}
\usepackage{xcolor}

\definecolor{promptbg}{gray}{0.95}
\lstdefinestyle{promptstyle}{
  basicstyle=\ttfamily\fontsize{6.5pt}{8pt}\selectfont,
  backgroundcolor=\color{promptbg},
  frame=single,
  framerule=0.5pt,
  rulecolor=\color{black!70},
  breaklines=true,
  breakatwhitespace=false,
  keepspaces=true,
  showstringspaces=false,
  tabsize=2,
  xleftmargin=1em,
  xrightmargin=1em,
  aboveskip=0.5em,
  belowskip=1em,
  breakindent=0pt,         
  breakautoindent=false,   
}

\title{Improving LLM's Attachment to External Knowledge In Dialogue Generation Tasks Through Entity Anonymization}

\author{
  Hadi Sheikhi, Chenyang Huang, Osmar R. Zaïane \\
  Dept. of Computing Science, University of Alberta \\
  Alberta Machine Intelligence Institute \\
  \texttt{\{hsheikhi,chenyangh,zaiane\}@ualberta.ca}
}

\begin{document}
\maketitle
\begin{abstract}
Knowledge graph-based dialogue generation (KG-DG) is a challenging task requiring models to effectively incorporate external knowledge into conversational responses. While large language models (LLMs) have achieved impressive results across various NLP tasks, their ability to utilize external knowledge in KG-DG remains under-explored. We observe that LLMs often rely on internal knowledge, leading to detachment from provided knowledge graphs, even when they are given a flawlessly retrieved knowledge graph. First, we introduce LLM-KAT, an evaluation procedure for measuring knowledge attachment in generated responses. Second, we propose a simple yet effective entity anonymization technique to encourage LLMs to better leverage external knowledge. Experiments on the OpenDialKG dataset demonstrate that our approach improves LLMs' attachment on external knowledge.\footnote{ \href{https://github.com/Hadishh/llm_attachment}{https://github.com/Hadishh/llm\_attachment}}
\end{abstract}

\section{Introduction}  

Knowledge Graph-based Dialogue Generation (KG-DG) aims to generate a response based on a retrieved subgraph, conditioned on the dialogue history \cite{moon-etal-2019-opendialkg}. As a highly informative data format, Knowledge Graphs (KGs) are considered to be an effective tool for dialogue generation systems \cite{han-etal-2015-exploiting, ijcai2018p0643}. Recent studies \cite{ji-etal-2023-rho, park-etal-2024-generative} employ small Pre-trained Language Models (PLMs) for response generation in KG-DG. However, carefully designed models are dataset-tailored and require further fine-tuning or pre-training.

Recently, Large Language Models (LLMs; \citealt{bai2023qwentechnicalreport, touvron2023llamaopenefficientfoundation, openai2024gpt4technicalreport}) have shown strong capabilities in conversational agents \cite{liao-etal-2023-proactive} and knowledge-graph grounding tasks \cite{agarwal-etal-2024-symkgqa, liu-etal-2025-superficial, yang-etal-2025-curiousllm}. However, their performance on KG-DG has not yet been fully explored. Our preliminary experiments show that LLMs outperform existing baselines even in a zero-shot setting. In particular, the generation fluency and context coherence are much improved with LLMs.

\begin{figure}[!t]
    \centering
    \includegraphics[scale=.5]{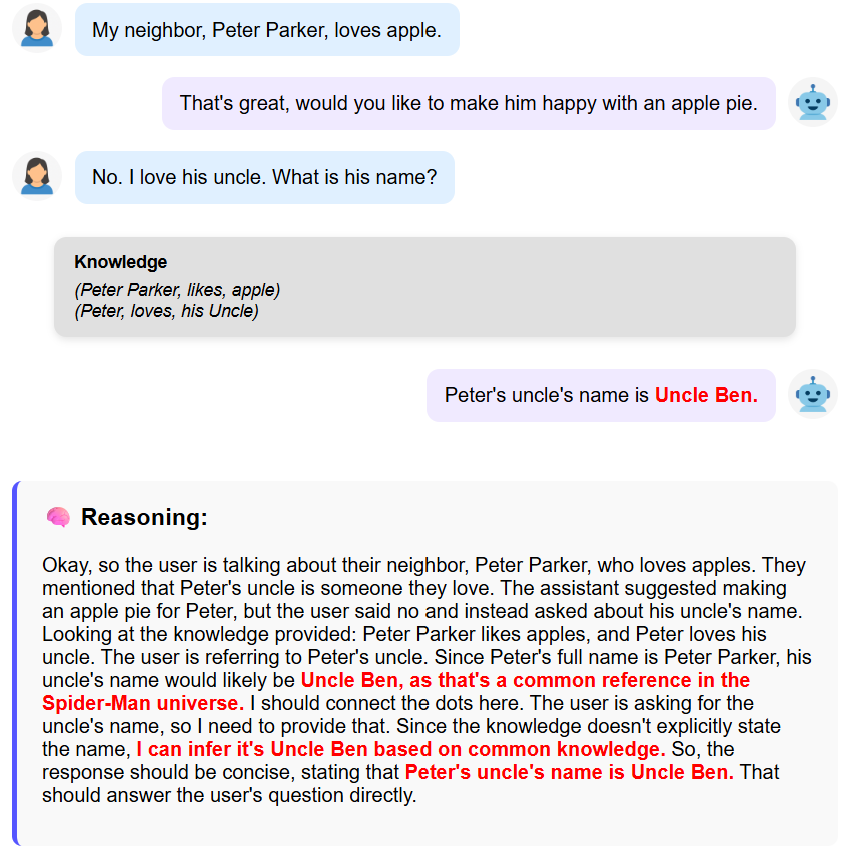}
    \caption{Example of an additional assumption made by the LLM, based on its internal parameters, when generating the dialogue response.}
    \label{fig:1}
\end{figure}

Nevertheless, we identify a common type of error by LLM. As seen in Figure \ref{fig:1}, LLMs tend to detach from an even perfectly retrieved knowledge, to complete the conversation, as they have access to the enormous internal knowledge obtained from the pretraining \cite{kadavath2022languagemodelsmostlyknow}. This detachment, however, often results in inappropriate responses.

In this work, we rigorously investigate the detachment of LLMs from context in the KG-DG task. First, we introduce a more reliable evaluation procedure using LLMs as a robust entity extractor to measure the detachment of knowledge. Specifically, we transform the evaluation into Question Answering (QA) format and prompt a strong LLM to extract the answers from the generated response. This approach enables us to systematically assess the extent to which LLMs utilize external knowledge versus relying on their internal knowledge.


Further, we propose a simple yet effective method to improve LLMs' attachment to any given knowledge. Specifically, we anonymize the entities (e.g., names of people and movies) in the dialogue history and the given knowledge with an identifier. In this way, the entities in conversation would not match the internal knowledge of LLM, forcing the reasoning of response generation to be attached to the given knowledge.

We conduct experiments on two versions of the OpenDialKG dataset \cite{moon-etal-2019-opendialkg}: the Normal and an Anonymized variant. Our findings show improvements on the attachment to the provided knowledge under anonymization, suggesting that anonymization mitigates the influence of internal knowledge in response generation. We also show that such anonymization does not significantly affect the quality of LLM's responses~\cite{pasch-cha-2025-balancing}. 
\section{Related Work}

\textbf{Knowledge Graph Dialogue Generation}. Various studies incorporate external knowledge graph into the dialogue generation \cite{han-etal-2015-exploiting, eric-etal-2017-key, madotto-etal-2018-mem2seq, ijcai2018p0643, moon-etal-2019-opendialkg, tuan-etal-2019-dykgchat, zhang-etal-2020-grounded, galetzka-etal-2021-space, zhou-etal-2021-earl, rony-etal-2022-dialokg}. For instance, SURGE \cite{kang2023knowledgegraphaugmentedlanguagemodels} applies Graph Neural Networks (GNN) to retrieve context-relevant subgraphs. RHO \cite{ji-etal-2023-rho} utilizes graph embedding methods to generate the responses relevant to dialogue history. \citet{park-etal-2024-generative} proposed DialogGSR, which employs a T5-small model \cite{10.5555/3455716.3455856} for KG-DG. However, none these works considered incorporating knowledge triplets and LLMs solely for KG-DG.

\textbf{LLMs and Knowledge Graphs}. In recent years, a large flow of research has attempted to integrate KGs with LLMs to combine their strengths in several domains, especially in Question Answering \cite{salnikov-etal-2023-large, sen-etal-2023-knowledge, Guo_2024, xu2025harnessinglargelanguagemodels}. Another line of work is focusing on reasoning on knowledge graphs to find the best answer \cite{fang-etal-2024-trace, luo2024graphconstrainedreasoningfaithfulreasoning, amayuelas2025groundingllmreasoningknowledge}. However, the process through which external knowledge is utilized by LLMs in KG-DG remains under-explored.

\textbf{Anonymization} techniques are widely recognized for preserving data privacy \cite{k-anonymity, l-diversity-2006}. Additionally, recent research has employed anonymization to investigate whether LLMs leverage provided textual contexts or not. For instance, \citet{longpre2022entitybasedknowledgeconflictsquestion} introduced an anonymized variant of the Natural Questions benchmark \cite{10.1162/tacl_a_00276}, wherein the original answer entity is replaced with an alternative name. Under such condition, one could assess whether an LLM is using its parametric memory or is grounded to context. Despite prior studies, our work adopts anonymization as a means to encourage LLMs to rely on external knowledge rather than their internal knowledge.
\section{Methodology}
In this section, we first present the prompting strategy employed for knowledge graph-based dialogue generation, followed by a detailed description of LLM-KAT, our proposed evaluation procedure. We then introduce the anonymization strategy used to mitigate detachment from context. We compare performance on both the Anonymized and Normal datasets using LLM-KAT as the primary evaluation framework and demonstrate the effectiveness of anonymization on attachment.\\
\textbf{Knowledge Graph Dialogue Generation.}
The primary objective of KG-DG is to generate an appropriate response given the dialogue history and its associated subgraph. The dialogue history contains turns between user and assistant, with the subgraph consisting of knowledge triplets for both the preceding conversation and the forthcoming response. We present each subgraph to the language model as a sequence of triplets in their canonical form $(e_1, r, e_2)$, where $e_1$ and $e_2$ denote the head and tail entities, respectively, and $r$ represents the relation linking them.
We prompt the LLM to generate a response conditioned on the dialogue history and associated subgraph. The response generation prompt is provided in Appendix~\ref{app:kgdg_prompt}.\\
\begin{table*}[ht]
  \centering
  \scriptsize
  \begin{tabular}{llccc cc}
    \toprule
    \multirow{2}{*}{\textbf{Model}} 
      & \multirow{2}{*}{\textbf{Dataset}} 
      & \multicolumn{2}{c}{\textbf{LLM-KAT}} 
      & \multicolumn{2}{c}{\textbf{UniEval}} \\
    \cmidrule(lr){3-4} \cmidrule(lr){5-6}
      &                          & \textbf{F1 Per Turn} & \textbf{F1 Per Session} 
      & \textbf{Naturalness} & \textbf{Coherence} \\
    \midrule
    DeepSeek-r1-7B   & Normal     & 76.04 & 77.16 & 92.36 & 98.59 \\
                     & Anonymized & 76.37 (\textbf{+0.33})  & 77.77 (\textbf{+0.61}) & 88.89 (\textbf{-3.47}) & 97.75 (\textbf{-0.84}) \\
    \midrule
    Qwen-7B           & Normal     & 82.38 & 83.55 & 92.89 & 98.85 \\
                     & Anonymized  & 83.86 (\textbf{+1.48}) & 84.78 (\textbf{+1.23}) & 90.12 (\textbf{-2.77}) & 98.42 (\textbf{-0.43}) \\
    \midrule
    DeepSeek-r1-14B  & Normal     & 84.84 & 86.24 & 93.71 & 98.37 \\
                     & Anonymized & 89.19 (\textbf{+4.35}) & 90.95 (\textbf{+4.71}) & 91.50 (\textbf{-2.21}) & 98.09 (\textbf{-0.28}) \\ 
    \midrule
    Qwen-14B          & Normal    & 83.87 & 84.73 & 93.01 & 99.32 \\ 
                     & Anonymized & 88.91 (\textbf{+5.04}) & 90.00 (\textbf{+5.27}) & 93.44 (\textbf{+0.43}) & 99.30 (\textbf{-0.02}) \\
    \midrule
    DeepSeek-r1-32B  & Normal     & 85.56 & 86.80 & 94.83 & 99.02 \\
                     & Anonymized & 89.75 (\textbf{+4.19}) & 91.46 (\textbf{+4.66}) & 92.90 (\textbf{-1.93}) & 98.84 (\textbf{-0.18}) \\
    \midrule
    Qwen-32B          & Normal    & 86.57 & 87.06 & 89.90 & 99.50 \\
                     & Anonymized & 90.14 (\textbf{+3.57}) & 91.13 (\textbf{+4.07}) & 88.54 (\textbf{-1.36}) & 99.44 (\textbf{-0.06}) \\
    \bottomrule
  \end{tabular}
  \caption{Performance of various LLMs under normal vs.\ anonymized dataset settings, evaluated on F1 (per-turn/session) for LLM-KAT, and UniEval metrics.}
  \label{tab:1}
\end{table*}
\textbf{Knowledge Attachment Test (KAT).}
\label{sec:kqa_llm}
We introduce an evaluation procedure (LLM-KAT) to measure the attachment of responses to their corresponding subgraphs. We transform the evaluation task into Question Answering (QA). Specifically, given a multi-turn conversation $C = \{T_1, T_2, \dots \}$ between a user and an assistant, along with a set of knowledge triplets $K$ pertinent to response generation, we treat last conversational turn as the input context and its associated triplet $(e_1, r, e_2)$ as the corresponding question.
In contrast to previous work \cite{kang2023knowledgegraphaugmentedlanguagemodels}, we leverage LLMs to answer the question: "Given the triplet $(e_1, r, X)$, which span of the input ($T_i$) fills $X$?" in an efficient way. With this approach, LLM extracts candidate entities as replacements for $X$.  We compare these candidate entities $e_2'$ with the ground truth $e_2$ using the SQuAD~\cite{rajpurkar-etal-2016-squad} F1-score to assess the attachment of each turn to the provided triplet. The model is instructed to generate candidates for 20 samples in a single response. The prompt template is provided in Appendix~\ref{app:llm_kqa}.\\
\textbf{Anonymization.}
We prompt the LLM to anonymize the entire context, following \citet{staab2025largelanguagemodelsadvanced}, who demonstrate that LLMs are expert anonymizers. Using in-context learning \cite{NEURIPS2020_1457c0d6}, we first instruct the model to generate a mapping table between entities in the context and their type, augmented with a sequential identifier (e.g., "Person1"). The model then regenerates the conversation, replacing all entity mentions with their anonymized forms. For example, "Robert Downey Jr." may be referred to as "RDJ", which rule-based systems often miss, while LLM-based methods can identify and anonymize such variants. The anonymization prompt is provided in Appendix~\ref{app:anon_prompt}.

\section{Experimental Setup}
\label{sec:exp}
We experiment several LLMs on the KG-DG task on two versions of the dataset. For each dialogue turn, we generate a response conditioned on the dialogue history and its corresponding knowledge triplet. The generated responses are then assessed using our proposed LLM-KAT metric, which quantifies contextual attachment. Further details of our experimental setup are presented below.
\subsection{Datasets}
We perform our experiments on two versions of the OpenDialKG dataset \cite{moon-etal-2019-opendialkg}: the original (Normal) and an anonymized variant. OpenDialKG comprises approximately 15K human-to-human conversations, of which 12K are annotated with knowledge triplets aligned to the dialogue history. The dialogues are originally collected and annotated against a large external KG, resulting in a per-dialogue subgraph. We segment these conversations into dialogue turns, yielding around 37K turns with their corresponding triplets. Additionally, to compare our evaluation procedure (LLM-KAT) with KQA \cite{kang2023knowledgegraphaugmentedlanguagemodels}, we create a synthetic dataset containing 20k turn-triplet pairs that are unanswerable by design. Specifically,  we use the constant string "TARGET\_ENTITY is not found in the database" as the base context. For each triplet of the form $(e_1, r, X)$, we replace the target entity with ground truth value of X. For example, in the triplet (GOT, written\_by, X), the resulting context becomes \textit{"George R.R. Martin is not found in the database"}. This context-question pair is designed such that the answer is IS\_IMPOSSIBLE. A robust evaluation procedure have the ability to maintain such question-context pairs.\\
\subsection{Evaluation Metrics}
We evaluate the attachment using LLM-KAT (Section~\ref{sec:kqa_llm}). Specifically, we assess how well the LLMs align with the provided knowledge triplets by computing the F1-score at both the turn level (micro average) and the session level (macro average).
To assess the impact of anonymization on response quality, we adopt the Coherence and Naturalness metrics introduced by UniEval~\cite{zhong-etal-2022-towards}. These metrics have demonstrated strong correlation with human judgments in their experiments.\\
\subsection{Baseline}
We compare DialogGSR~\cite{park-etal-2024-generative} with LLMs to demonstrate the relevance of LLM-based approaches for KG-DG. The evaluation is conducted on their test set consisting of 1,082 turns, of which 759 are mapped to the OpenDialKG dataset. Our comparison between DialogGSR and LLMs is performed on these aligned samples.
To evaluate the impact of anonymization on LLMs' sensitivity to attachment, we compare each model's performance on the Normal and Anonymized versions of the dataset. The Normal dataset serves as the baseline in each experiment.\\
\subsection{Settings}
We evaluate both standard and reasoning-oriented LLMs, focusing on Qwen2.5-Instruct\footnote{\href{https://huggingface.co/spaces/Qwen/Qwen2.5}{https://huggingface.co/spaces/Qwen/Qwen2.5}} \cite{bai2023qwentechnicalreport} and DeepSeek-R1\footnote{\href{https://huggingface.co/deepseek-ai/DeepSeek-R1}{https://huggingface.co/deepseek-ai/DeepSeek-R1}} \cite{deepseekai2025deepseekr1incentivizingreasoningcapability}. To analyze the impact of model scale, we consider variants with 7B, 14B, and 32B parameters. DeepSeek-R1-32B is further utilized for LLM-KAT extraction. We employed vLLM~\cite{kwon2023efficient} library for all of our experiments.
For the anonymization process, we utilize QwQ\footnote{\href{https://huggingface.co/Qwen/QwQ-32B}{https://huggingface.co/Qwen/QwQ-32B}} \cite{qwq32b} as the perfect anonymizer for our controlled experiments. We did not further investigate anonymization quality as it is beyond the scope of this work.
\section{Analysis}
\label{sec:analysis}
\textbf{Effect of Anonymization on Contextual Attachment.} Table~\ref{tab:1} shows that all models demonstrate increased attachment on dialogue context when anonymization is applied. This finding indicates that anonymization effectively limits the use of internal knowledge by LLMs, thereby enhancing contextual attachment and reducing the likelihood of generating irrelevant responses.\\
\begin{table}[h]
  \centering
  \scriptsize
    \begin{tabular}{lcccc}
      \toprule
      \textbf{Model} 
        & \textbf{F1/Turn} 
        & \textbf{Coherence} \\
      \midrule
DialogGSR \cite{park-etal-2024-generative} & 55.59 & 87.08 \\
DeepSeek-7B                  & 78.19  & 98.83 \\
Qwen-7B                      & \textbf{85.16}  & \textbf{99.04} \\
      \bottomrule
    \end{tabular}
\caption{Comparing LLMs with small PLMs}
\label{tab:2}
\end{table}
\textbf{LLMs on KG-DG}. As shown in Table \ref{tab:2}, LLMs outperform smaller PLMs on the KG-DG task. This result highlights the effectiveness of LLMs in zero-shot settings and supports our motivation to leverage LLMs for KG-DG without additional fine-tuning.
\begin{table}[!h]
  \centering
  \scriptsize
  \begin{tabular}{@{}lcc@{}}
    \toprule
    \textbf{Model}   & \textbf{IMP} & \textbf{FP}  \\
    \midrule
    KQA \cite{kang2023knowledgegraphaugmentedlanguagemodels}        & N/A               & 100        \\
    LLM-KAT          & \textbf{78.92}             & \textbf{21.08}        \\
    \bottomrule
  \end{tabular}
  \caption{Performance of LLM-KAT and KQA on impossible to answer syntethic dataset.}
  \label{tab:3}
\end{table}\\
\textbf{LLM-KAT Performance.} Unlike KQA metric~\cite{kang2023knowledgegraphaugmentedlanguagemodels}, LLM-KAT identifies cases where the question is unanswerable in the given context. As shown in Table~\ref{tab:3}, using BERT for span extraction results in predicting the target entity in a context unrelated to the question triplet, leading to inflated scores under conventional KQA evaluation. In contrast, LLM-KAT is able to flag a substantial proportion of synthetic context-triplet pairs as unanswerable (79\% predicted as impossible), and report F1 scores only on the remaining answerable cases. This approach reduces the risk of False Positive (FP) predictions and provides a more accurate assessment of model performance.
\\
\textbf{Prompt Engineering.} Observations in Table~\ref{tab:prompt_eng} indicate that even a more detailed prompt (see Appendix~\ref{app:kgdg_prompt}), which includes explicit instructions to generate appropriate responses, can still benefit from anonymization. Furthermore, our results show that as model size increases, the impact of prompt detail on contextual attachment decreases. Specifically, the 7B model exhibits a notable improvement in contextual attachment when given detailed prompts on the Normal dataset, whereas the 32B model shows only a modest gain. We do not further explore the few-shot setting, as prior work has shown it can degrade the performance of reasoning LLMs~\cite{deepseekai2025deepseekr1incentivizingreasoningcapability}.
\begin{table}[h!]
\centering
\scriptsize
\begin{adjustbox}{width=\columnwidth}
\begin{tabular}{lllccll}
\toprule
\textbf{Model} & \textbf{Prompt} & \textbf{Dataset} & \multicolumn{2}{c}{\textbf{LLM-KAT}} \\
\cmidrule(lr){4-5}
& & & \textbf{F1/Turn} & \textbf{F1/Session} \\
\midrule
\multirow{4}{*}{DeepSeek-7B} 
& Default & Normal      & 76.04 & 77.16 \\
&         & Anonymized  & 76.37 & 77.77 \\
\cmidrule(lr){2-5}
& Detailed & Normal     & 79.11 & 81.35 \\
&         & Anonymized  & 80.45 & 82.85 \\
\midrule
\multirow{4}{*}{DeepSeek-14B} 
& Default & Normal      & 84.84 & 86.24 \\
&         & Anonymized  & 89.19 & 90.95 \\
\cmidrule(lr){2-5}
& Detailed & Normal     & 87.04 & 89.02 \\
&         & Anonymized  & 89.89 & 91.78 \\
\midrule
\multirow{4}{*}{DeepSeek-32B} 
& Default & Normal      & 85.56 & 86.80 \\
&         & Anonymized  & 89.75 & 91.46 \\
\cmidrule(lr){2-5}
& Detailed & Normal     & 87.69 & 89.38 \\
&         & Anonymized  & \textbf{90.73} & \textbf{92.61}  \\
\bottomrule
\end{tabular}
\end{adjustbox}
\caption{LLM-KAT scores across different DeepSeek models, prompts, and datasets.}
\label{tab:prompt_eng}
\end{table}\\
\textbf{Partial Anonymization.} We curate a half anonymized dataset by merging 50\% of the normal and anonymized datasets (see Appendix~\ref{app:half_anon} for details). Our results on half anonymized dataset demonstrate that contextual attachment gradually increases from the normal to half-anonymized and fully anonymized datasets for LLMs with reasoning capabilities. This trend supports the conclusion that anonymization is a primary factor driving LLMs to rely more on dialogue context. The results for these experiments are provided in Appendix~\ref{app:half_anon}.\\
\textbf{Impact of Anonymization on Response Quality.} The effect of anonymization on response quality is minimal, as evidenced by the Coherence and Naturalness scores across datasets in Table~\ref{tab:1}. This finding aligns with the observations of \citet{pasch-cha-2025-balancing}, indicating that anonymization does not substantially degrade response quality. In fact, this performance degradation is expected, as our "anonymization" approach effectively applies control over the LLM's generation. In contrast, an unconstrained alternative would naturally achieve higher scores, which are themselves assigned by another language model~\cite{zhong-etal-2022-towards}.  Furthermore, Table~\ref{tab:5} shows that the quality gap between anonymized and non-anonymized responses narrows as model size increases, suggesting that larger LLMs maintain high response quality regardless of anonymization.
\begin{table}[!h]
  \centering
  \scriptsize
  \begin{tabular}{@{}lcc@{}}
    \toprule
    \textbf{Model}   & \textbf{C-Drop (\%)} & \textbf{N-Drop (\%)}\\
    \midrule
    DeepSeek-7B            & 0.84 & 3.47  \\
    DeepSeek-14B           & 0.28 & 2.21 \\
    DeepSeek-32B           & \textbf{0.18} & \textbf{1.93} \\
    \bottomrule
  \end{tabular}
  \caption{The effect of anonymization on Coherence (C) and Naturalness (N) drop for reasoning LLMs.}
  \label{tab:5}
\end{table}\\
\textbf{Qualitative Analysis.} To further demonstrate the effectiveness of our anonymization approach, we conduct a qualitative analysis of the models' responses. A strong LLM (DeepSeek-R1-32B) is employed as a judge to select the better response between the normal and anonymized variants. We use a multiple-choice prompt (A or B) for the comparison (see Appendix~\ref{app:qual_prompt}). To mitigate positional bias, the normal and anonymized responses are randomly assigned to options A or B for each evaluation instance with equal (50\%) probability.
\begin{table}[htb]
\centering
\scriptsize
    \begin{tabular}{lccc}
        \hline
        \textbf{Model} & \textbf{Normal (\%)} & \textbf{Tie (\%)} & \textbf{Anonymized (\%)} \\
        \hline
        DeepSeek-7B  & 36.27 & 22.29 & \textbf{41.44} \\
        DeepSeek-14B & 27.36 & \textbf{38.66} & 33.98 \\
        DeepSeek-32B & 29.01 & 34.18 & \textbf{36.81} \\
        \hline
    \end{tabular}
\caption{Qualitative Analysis using LLM-as-judge}
\label{tab:qualitative_analysis}
\end{table}\\
As shown in Table~\ref{tab:qualitative_analysis}, the anonymized responses were favored by the judge in two model sizes (7B and 32B), and the 14B model exhibited a higher proportion of ties. This indicates that the anonymization approach is effective in improving the overall quality of generated responses.\\
\textbf{KG-DG with stronger LLM. } To further validate the effectiveness of our approach, we conduct additional experiments using a stronger LLM, QwQ, on KG-DG task. Based on the results in Table~\ref{tab:qwq_ds}, QwQ benefits from anonymization in terms of contextual attachment, similar to the results observed with DeepSeek-R1. Surprisingly, despite DeepSeek-R1 models, we have no naturalness and coherence drop on QwQ-32B. This indicates that QwQ is more robust to anonymization, likely due to its enhanced capabilities.

\begin{table}[h!]
\centering
\begin{adjustbox}{width=\linewidth}
\begin{tabular}{llcccc}
\toprule
\textbf{Model} & \textbf{Dataset} & \multicolumn{2}{c}{\textbf{LLM-KAT}} & \multicolumn{2}{c}{\textbf{UniEval}} \\
\cmidrule(lr){3-4} \cmidrule(lr){5-6}
& & \textbf{F1/Turn} & \textbf{F1/Session} & \textbf{Naturalness} & \textbf{Coherence} \\
\midrule
\multirow{2}{*}{QwQ} 
& Normal      & 85.21 & 86.62 & 92.48 & 97.98 \\
& Anonymized  & 91.2 (\textbf{+5.99}) & 92.81 (\textbf{+6.19}) & 93.62 (\textbf{+1.14}) & 98.7 (\textbf{+0.72}) \\
\midrule
\multirow{2}{*}{DeepSeek-r1-32B}
& Normal      & 85.56 & 86.80 & 94.83 & 99.02 \\
& Anonymized  & 89.75 (\textbf{+4.19}) & 91.46 (\textbf{+4.66}) & 92.90 (\textbf{-1.93}) & 98.84 (\textbf{-0.18}) \\
\bottomrule
\end{tabular}
\end{adjustbox}
\caption{Comparison of QwQ and DeepSeek-r1-32B on Normal vs Anonymized datasets using LLM-KAT and UniEval metrics. Anonymization gains are bolded.}
\label{tab:qwq_ds}
\end{table}

\section{Conclusion}
We investigate the detachment of large language models (LLMs) from context in knowledge graph-based dialogue generation. We introduce a more robust evaluation procedure to quantify this detachment and propose data anonymization prior to inference to enhance contextual attachment. Our findings on 6 types of LLMs with two different prompting styles demonstrate that anonymization consistently improves attachment to the provided knowledge by constraining LLMs from leveraging internal knowledge. These findings suggest that our approach generalizes across model types and prompt designs.

\section*{Limitations}
For research purposes, this work relies on a heavy model (QwQ-32B) as an expert anonymizer, which is costly and requires significant computational resources. However, our studies on the partially anonymized dataset demonstrate that this technique is effective even with a weaker anonymizer by sacrificing some performance. 

Furthermore, our experiments were conducted solely on the OpenDialKG dataset, as more complex knowledge graph-based dialogue generation datasets are not currently available in literature.

\section*{Acknowledgments}
This work was supported by Alberta Machine Intelligence Institute (Amii).
\bibliography{anthology,custom}
\appendix

 
\section{Prompts}
\label{app:prompts}
\subsection{KG-DG Prompts}
Prompts for dialogue generation for DeepSeek is in Tables \ref{tab:dialog_generations} and \ref{tab:detailed_prompt}. We utilized the same prompt for Qwen, removing \textit{<think>} token. 
\label{app:kgdg_prompt}
\begin{table}[ht]
  \centering
  \begin{minipage}{\linewidth}
    \lstset{style=promptstyle}
    \begin{lstlisting}
You are an expert dialogue agent. Use the provided conversation history and external knowledge (as triplets) to generate a precise, fact-based reply. The response should not be longer than 2-3 sentences. Remember to stick to the knowledge.

History:
{history}

Knowledge:
{external_kg}

<think>
    \end{lstlisting}
  \end{minipage}
  \caption{Prompt for simple dialogue generation for LLMs.}
  \label{tab:dialog_generations}
\end{table}\\
\begin{table}[ht]
  \centering
  \begin{minipage}{\linewidth}
    \lstset{style=promptstyle}
    \begin{lstlisting}
You are a concise dialogue agent. Your task is to generate a short, precise, and fact-based response grounded strictly in the provided external knowledge, which is given in the form of structured triplets. Use the conversation history to understand the context, but only use facts that are directly supported by the external knowledge when forming your reply. Do not infer any information not explicitly present in the external knowledge.

Guidelines:

- Limit your response to 2-3 sentences.
- Use clear and factual language.
- Do not invent or speculate beyond the given knowledge.
- You may rephrase or combine knowledge triplets for naturalness, but do not introduce new facts.

Conversation History:
{history}

External Knowledge (Triplets):
{external_kg}

<think>
    \end{lstlisting}
  \end{minipage}
  \caption{A more detailed prompt for dialogue generation, emphasizing concise and fact-based responses grounded in external knowledge.}
  \label{tab:detailed_prompt}
\end{table}

\subsection{Anonymization Prompt}
\label{app:anon_prompt}
Our in-context learning prompt for anonymization is provided in Table \ref{tab:anonym_prompt}.

\subsection{Qualitative Analysis Prompt}
\label{app:qual_prompt}
Our qualitative analysis prompt is provided in Table~\ref{tab:qualitative_analysis_prompt}.
\subsection{LLM-KAT Entity Extraction}
\label{app:llm_kqa}
\begin{table}[ht]
  \centering
  \begin{minipage}{\linewidth}
    \lstset{style=promptstyle}
    \begin{lstlisting}
Your task is to extract, similar to the SQuAD V2 dataset, from the provided context that fills triplets. The triplets are of format (Subject, predicate, Object) in which either Subject or Object is missing. I use X instead of missing Subject or Object in the triplet, you need to extract the appropriate values for the X from the context. Each extracted answer must be an exact substring from the context. If there are multiple valid answers, separate them with @@. If no answer exists in the context, output IS_IMPOSSIBLE. 


Input Structure: The input is given in the following format: ID|||context|||triplet
There are 20 instances that you should response to. 

Output Format: The final output should follow this format: ID|||context|||triplet|||answer_1@@answer_2@@answer_3 Replace answer_1@@answer_2@@answer_3 with the actual answers found, or IS_IMPOSSIBLE if none are found. You should output 20 lines.

Input:
{samples}

<think>
    \end{lstlisting}
  \end{minipage}
  \caption{Prompt for LLM-KAT entity extraction}
  \label{tab:llm_kqa}
\end{table}

Our prompt for LLM-KAT is provided in Table \ref{tab:llm_kqa}.\\
\begin{table}[h]
  \centering
  \resizebox{0.5\textwidth}{!}{%
  \begin{tabular}{l c c c}
    \toprule
    Task                   & GPU-Type & \#GPUs & Time (H) \\
    \midrule
    Dialogue Generation    & I40s     & 4      & 2.322    \\
    LLM-KAT                & I40s     & 4      & 3.759    \\
    UniEval                & I40s     & 1      & 0.8      \\
    Anonymization          & I40s     & 4      & 12.021    \\
    \bottomrule
  \end{tabular}
  }
  \caption{Wall-on-clock time for experiments.}
  \label{tab:gpu_runtimes}
\end{table}
\section{Half Anonymized Dataset}
\label{app:half_anon}
We merge two anonymized and normal dataset by sampling 50\% from each dataset. By doing so, we have a half anonymized dataset. The results for reasoning LLMs on this dataset are reported in Table \ref{tab:half_anon_results}.

\begin{table}[htb]
  \centering
  \resizebox{0.4\textwidth}{!}{%
  \begin{tabular}{lcc}
  \toprule
  \textbf{Hyperparameter} & \textbf{DeepSeek} & \textbf{Qwen} \\ \midrule
  Max new Tokens & 16384 & 256 \\
  Top K & \multicolumn{2}{c}{10} \\
  Top P & \multicolumn{2}{c}{0.95} \\
  Temperature & \multicolumn{2}{c}{0.6} \\
  Batch Size & \multicolumn{2}{c}{128} \\ \bottomrule
  \end{tabular}
  }
  \caption{VLLM hyperparameters for Dialogue Generation.}
  \label{tab:hyperparameters}
  \end{table}

  \begin{table}[htb]
  \centering
  \resizebox{0.25\textwidth}{!}{%
  \begin{tabular}{lc}
  \toprule
  \textbf{Hyperparameter} & \textbf{Value} \\ \midrule
  Seed & 42 \\
  Max new Tokens & 16384 \\
  Top K & 10 \\
  Top P & 0.95 \\
  Temperature & 0.6 \\
  Batch Size & 128 \\ \bottomrule
  \end{tabular}
  }
  \caption{VLLM hyperparameters for LLM-KAT.}
  \label{tab:hyperparameters_llm_kat}
  \end{table}

\begin{table*}[ht]
  \centering
  \small
  \begin{tabular}{llccc cc}
    \toprule
    \multirow{2}{*}{\textbf{Model}} 
      & \multirow{2}{*}{\textbf{Dataset}} 
      & \multicolumn{2}{c}{\textbf{LLM-KAT}} 
      & \multicolumn{2}{c}{\textbf{UniEval}} \\
    \cmidrule(lr){3-4} \cmidrule(lr){5-6}
      &                         & \textbf{F1 Per Turn} 
                                & \textbf{F1 Per Session} 
      & \textbf{Naturalness} 
      & \textbf{Coherence} \\
    \midrule
    \multirow{3}{*}{DeepSeek-r1-7B}
      & Normal           & 76.04 & 77.16 & 92.36 & 98.59 \\
      & Half Anonymized  & 75.31 & 76.53 & 90.73 & 98.11 \\
      & Anonymized       & 76.37 & 77.77 & 88.89 & 97.75 \\
    \midrule
    \multirow{3}{*}{DeepSeek-r1-14B}
      & Normal           & 84.84 & 86.24 & 93.71 & 98.37 \\
      & Half Anonymized  & 86.15 & 87.61 & 92.66 & 98.25 \\
      & Anonymized       & 89.19 & 90.95 & 91.50 & 98.09 \\
    \midrule
    \multirow{3}{*}{DeepSeek-r1-32B}
      & Normal           & 85.56 & 86.80 & 94.83 & 99.02 \\
      & Half Anonymized  & 87.24 & 88.65 & 93.90 & 98.92 \\
      & Anonymized       & 89.75 & 91.46 & 92.90 & 98.84 \\
    \bottomrule
  \end{tabular}
  \caption{DeepSeek performance under normal, half-anonymized, and fully anonymized input on LLM-KAT and UniEval.}
  \label{tab:half_anon_results}
\end{table*}
\section{Case Study}
\begin{figure*}[!ht]
  \centering
  \begin{subfigure}[b]{0.48\textwidth}
    \centering
    \includegraphics[width=\linewidth]{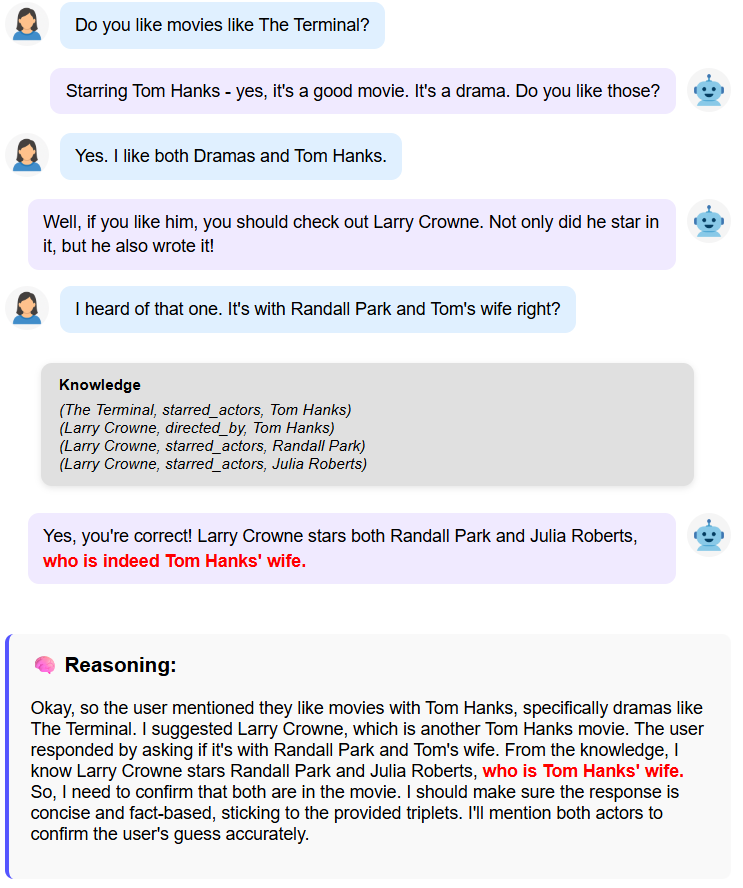}
    \caption{Normal dataset input.}
    \label{fig:case_normal}
  \end{subfigure}%
  \hfill
  \begin{subfigure}[b]{0.48\textwidth}
    \centering
    \includegraphics[width=\linewidth]{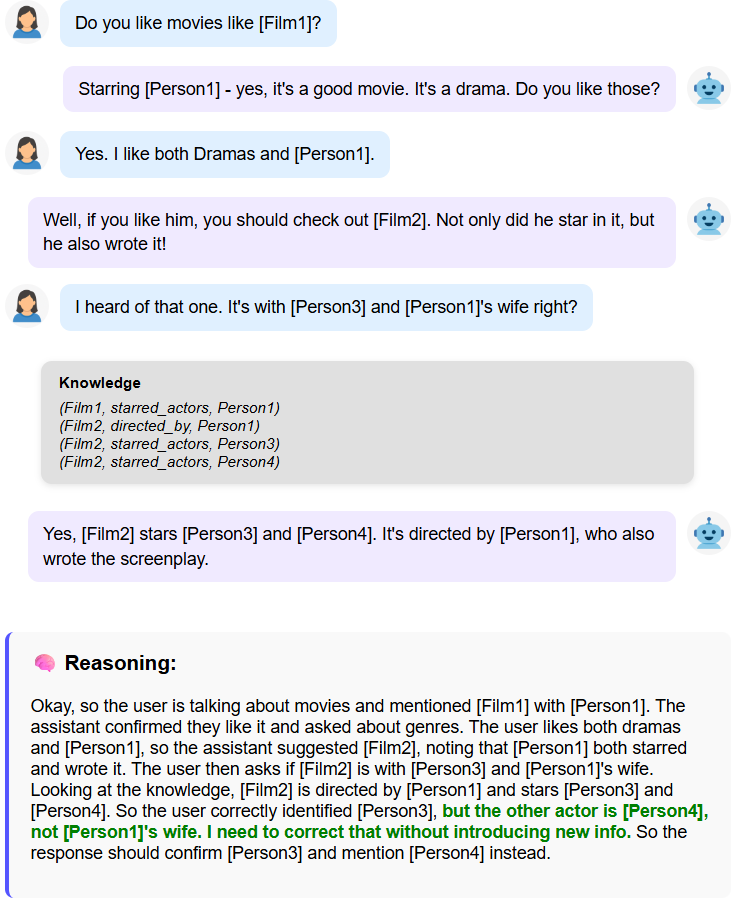}
    \caption{Anonymized dataset input.}
    \label{fig:case_anon}
  \end{subfigure}
  \caption{Case studies comparing (a) the normal dataset input and (b) the anonymized dataset input.}
  \label{fig:2}
\end{figure*}

As shown in Figure \ref{fig:case_normal}, the model made an assumption that Julia Roberts is the spouse of Tom Hanks. On the other hand, Figure \ref{fig:case_anon} shows a correction in reasoning while processing Tom's wife and does not detach from the given knowledge. This case shows the effect of anonymization on the response generation that it prevents the model from misinformation generation.

\begin{table*}[!htb]
  \centering
  
  \begin{minipage}{\textwidth}
    \lstset{style=promptstyle}
    \begin{lstlisting}
You are provided with a dialogue containing references to teams, people, locations, and literary influences, along with a set of related knowledge triplets. Your task is to anonymize the dialogue and the knowledge by replacing all real-world names with anonymized entity identifiers that include both a type (e.g., Team, Person, Place, LiteraryInfluence) and an ID. Be sure to create a clear mapping table, update all occurrences in the dialogue, and replace entries in the knowledge triplets accordingly.

Instructions:
1. Mapping Table:
   - First, scan the dialogue and the knowledge triplets to identify all unique entities (e.g., team names, person names, places, literary influences).
   - For each entity, define:
     - An anonymized ID. For example:
       - Films: F1, F2, ...
       - Teams: T1, T2, ...
       - People: P1, P2, ...
       - Places: L1, L2, ...
       - Literary Influences or Authors: LI1, LI2, ...
     - Its type (Team, Person, Place, LiteraryInfluence, Film, etc).
   - Create a mapping table that shows each entity's original name, its anonymized ID, and its type.

2. Anonymize the Dialogue:
   - Replace every occurrence of a real-world entity in the dialogue with its corresponding anonymized ID.
   - Ensure that informal or abbreviated references (if any) are also mapped correctly.

3. Anonymize the Knowledge Triplets:
   - For each knowledge triplet, replace each entity with its corresponding anonymized ID based on the mapping table.

4. Output Format:
   - Mapping Table: Present as a clear table with columns for ID, Original Entity, and Type.
   - Anonymized Dialogue: Provide the full dialogue with entity mentions replaced by their anonymized IDs.
   - Anonymized Knowledge Triplets: List the anonymized triplets exactly as in the original input, but with IDs in place of the original entity names.

Example In-Context (from a previous task):

Mapping Table Example:
| **ID** | **Original Entity**                   | **Type**  |
|--------|---------------------------------------|-----------|
| F1     | Iron Man                              | Film      |
| P1     | Robert Downey Jr.                     | Person    |
| F2     | Zodiac (Crime Fiction Film)           | Film      |
| P2     | Jake Gyllenhaal                       | Person    |
| F3     | End of Watch                          | Film      |
| P3     | David Ayer                            | Person    |
| G1     | Thriller                              | Genre     |
| G2     | Crime Fiction                         | Genre     |

Anonymized Dialogue Example:
- S1: Do you like [F1]?
- S2: Sure do! [P1] is a favorite.
- S1: Yes, I like him too. Did you know he also was in [F2], a [G2] film?
- S2: I like [G2]! Didn't know [P1] was in there. [P2] starred as well.
- S1: So he did? He also starred in [F3]. Have you ever seen that movie?
- S2: Yes, I have! I like films directed by [P3]. How about you?
- S1: I have not. What genre is [F3]?
- S2: It's a [G1] and [G2] film as well.
- S1: I will make sure to check it out. I like [G1] films. Thank you!
- S2: Welcome!

Anonymized Knowledge Triplets Example:
1. ["[F1]", "starred_actors", "[P1]"]
2. ["[F2]", "starred_actors", "[P1]"]
3. ["[F2]", "starred_actors", "[P2]"]
4. ["[F3]", "~starred_actors", "[P2]"]
5. ["[F3]", "written_by", "[P3]"]
6. ["[F3]", "has_genre", "[G1]"]

Now, please perform the anonymization for the following query:

------------------------------
Dialogue:
{history}

Knowledge:
{external_kg}
------------------------------

Ensure that your final output includes:
- A Mapping Table with all identified entities, their anonymized IDs, and their types.
- The complete dialogue with anonymized entities.
- The knowledge triplets with all entities replaced by their anonymized IDs.

<think>
    \end{lstlisting}
  \end{minipage}
  \caption{Anonymization Prompt}
  \label{tab:anonym_prompt}
\end{table*}
\begin{table*}[!htb]
  \centering
  
  \begin{minipage}{\textwidth}
    \lstset{style=promptstyle}
    \begin{lstlisting}
You are a high-precision quality assessment agent. Your task is to evaluate two candidate responses (Option A and Option B) and determine which one is the superior choice--or whether both are equally valid--based on the conversation history and structured external knowledge provided as triplets.

Follow these steps carefully:
    1. Understand the Context: Read the full conversation history to understand the dialogue intent, user queries, and tone.
    2. Analyze the Knowledge: Examine the knowledge triplets, each in the format (subject, predicate, object). Use them to fact-check and assess alignment with each option.
    3. Evaluate Each Option:
        - Check for factual correctness based on the triplets.
        - Make sure of the point that the response does not utilize any additional information/assumptions that is not provided in the context.
    4. Make a Decision:
        - Choose A if Option A is better in using knowledge triplets.
        - Choose B if Option B is better in using knowledge triplets.
        - Choose Both if both are equally valid in using knowledge triplets.

Only output one of the following as your final decision: A, B, or Both. Just provide the final decision.

Conversation History:
{history}

Knowledge Triplets:
{knowledge_triplets}

Candidate Responses:
A: {A}

B: {B}
<think>
    \end{lstlisting}
  \end{minipage}
  \caption{Qualitative Analysis Prompt}
  \label{tab:qualitative_analysis_prompt}
\end{table*}

\end{document}